\documentclass[10pt,twocolumn,letterpaper]{article}

\usepackage{iccv}              

\usepackage{graphicx}
\usepackage{amsmath}
\usepackage{amssymb}
\usepackage{booktabs}
\usepackage{multirow}
\usepackage{comment}
\usepackage{caption}
\usepackage{wrapfig}
\usepackage{enumitem}
\usepackage{mathtools}
\usepackage{array}
\usepackage{color}
\usepackage{algorithm}
\usepackage{colortbl}
\definecolor{greyline}{rgb}{0.105,0.410,0.113}
\newcommand{\squishlist}{
	\begin{list}{$\bullet$}
		{ \setlength{\itemsep}{0pt}
			\setlength{\parsep}{1pt}
			\setlength{\topsep}{1pt}
			\setlength{\partopsep}{0pt}
			\setlength{\leftmargin}{1.5em}
			\setlength{\labelwidth}{1em}
			\setlength{\labelsep}{0.5em} } }
\newcommand{\squishend}{\end{list} 
}

\usepackage[toc,page]{appendix}

\definecolor{iccvblue}{rgb}{0.21,0.49,0.74}
\usepackage[pagebackref,breaklinks,colorlinks,allcolors=iccvblue]{hyperref}

\def\paperID{xxx} 
\def\confName{ICCV}
\def\confYear{2025}

\title{AIM: Adaptive Inference of Multi-Modal LLMs via Token Merging and Pruning}

\author{Yiwu Zhong$^{1}$, Zhuoming Liu$^{2}$, Yin Li$^{2}$, Liwei Wang$^{1}$\thanks{Corresponding author.}\\
$^1$The Chinese University of Hong Kong, $^2$University of Wisconsin-Madison\\
}

\begin{document}
\maketitle

\begin{abstract}
Large language models (LLMs) have enabled the creation of multi-modal LLMs that exhibit strong comprehension of visual data such as images and videos. However, these models usually rely on extensive visual tokens from visual encoders, leading to high computational demands, which limits their applicability in resource-constrained environments and for long-context tasks. 
In this work, we propose a training-free adaptive inference method for multi-modal LLMs that can accommodate a broad range of efficiency requirements with a minimum performance drop. 
Our method consists of a) iterative token merging based on embedding similarity before LLMs, and b) progressive token pruning within LLM layers based on multi-modal importance. With a minimalist design, our method can be applied to both video and image LLMs.
Extensive experiments on diverse video and image benchmarks demonstrate that our method substantially reduces computation load (\eg, a \textbf{7-fold} reduction in FLOPs) while preserving the performance of video and image LLMs. Further, at a similar computational cost, our method outperforms the state-of-the-art methods in long video understanding (\eg, \textbf{+4.6} on MLVU). Additionally, our in-depth analysis provides insights into token redundancy and LLM layer behaviors, offering guidance for future research in designing efficient multi-modal LLMs. Our code is available at \url{https://github.com/LaVi-Lab/AIM}.

\end{abstract}

\section{Introduction}
\label{sec:intro}
Large language models (LLMs)~\cite{ouyang2022training, zhang2022opt, touvron2023llama, vicuna2023, yang2024qwen2technicalreport} have been recently adapted for visual understanding, fostering the developments in both image~\cite{liu2023improvedllava,dai2023instructblip, zhu2023minigpt4, chen2023shikra, huang2024language} and video LLMs~\cite{Maaz2023VideoChatGPT, lin2023video, li2023mvbench, xu2024pllava, zhang2024llavanext-video}. 
However, these multi-modal LLMs (MLLMs) usually rely on a large number of visual tokens generated by visual encoders~\cite{radford2021learning,zhai2023sigmoid}, especially for video data where token counts can reach thousands per video.
Such high token number requires extensive computational resources for both training and inference, restricting their use in real-world applications (\eg, real-time processing on mobile devices). 
Further, as the number of video frames increases, the total number of tokens also grows, limiting the model's capacity to handle dense video frames. This often results in the loss of crucial temporal information, which is essential for comprehending long videos~\cite{zhou2024mlvucomprehensivebenchmarkmultitask,zhang2024long,shang2024interpolatingvideollmslongersequencelmms}.

\begin{figure}[t]
    \centering
    \includegraphics[width=0.95\linewidth]{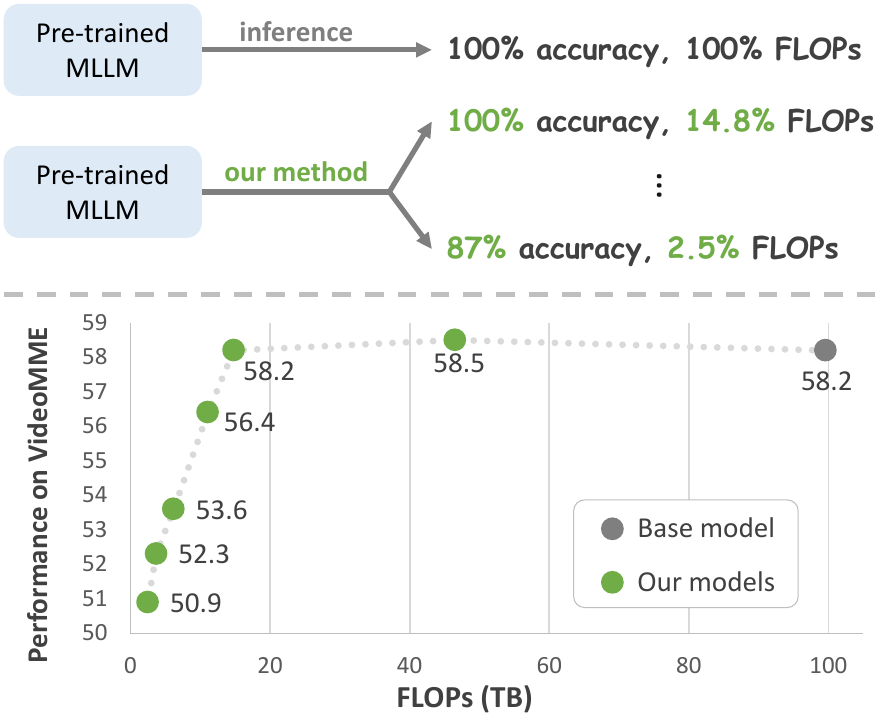}
    \vspace{-8pt}
    \caption{\textbf{Key idea and finding}. Our training-free method enables adaptive inference of pre-trained multi-modal LLMs, supporting a wide range of computational conditions. In comparison to the base pre-trained model, our method substantially reduces FLOPs, without or with a manageable performance drop.}
    \label{fig:teaser}
\end{figure}

To bridge the gap, we propose leveraging the inherent redundancy present in visual data to develop \textit{adaptive inference} for multi-modal LLMs. Adaptive inference dynamically adjusts a model's computational load during inference based on contextual factors~\cite{han2021dynamic}, \eg, computational constraints, image content, or desired accuracy levels. Since retaining all visual tokens during inference is often unnecessary, our key insight is to strategically select these tokens throughout the inference process. This approach allows for controlling the amount of computation necessary for inference, so as to optimize the model's efficiency while preserving its accuracy.

In this work, we introduce a training-free method of adaptive inference tailored for multi-modal LLMs, consisting of token merging based on visual similarity and token pruning based on multi-modal importance. Specifically, the visual encoder transforms the input visual data into visual tokens, which are iteratively merged based on their embedding similarities. The merged tokens are then fed into the LLM, where the tokens considered less important for multi-modal reasoning are progressively pruned at each layer, resulting in a gradual reduction of tokens. By adjusting the key parameters in the token merging and token pruning process, our method enables adaptive inference that achieves various levels of computation reduction with manageable performance loss, shown in Figure~\ref{fig:teaser}. With this capability, we can select a configuration of token merging and pruning so that the model meets the resource constraint while the performance is optimized.

Additionally, during the development of our method, we have several findings that can be useful for designing efficient multi-modal LLMs in the future. 
\textbf{First}, the full set of video tokens is often unnecessary, as only 25\% visual tokens fed into LLM can maintain close performance. 
\textbf{Second}, with fewer tokens per frame, LLMs can process a larger number of frames, thereby addressing information loss and improving performance, especially for long video understanding.
\textbf{Third}, we observe that pruning text tokens across LLM layers or removing visual tokens in the early LLM layers significantly impacts performance. In contrast, pruning a large proportion of visual tokens in the later layers can maintain performance. These observations suggest that multi-modal LLMs focus on cross-modal fusion in the earlier layers while prioritizing text tokens in later layers. 

We conduct extensive experiments on diverse video benchmarks and image benchmarks, with base models as LLaVA-OneVision~\cite{shang2024llava} and LLaVA-1.5~\cite{liu2023improvedllava}, respectively. Without additional fine-tuning, our method achieves a substantial reduction in computational demands (\eg, decreasing FLOPs and prefill time by a factor of \textbf{6.8} and \textbf{8.0}, respectively) while retaining performance close to the base video LLM and image LLM. 
Moreover, given the same computation resources, our method allows for involving more frames as inputs and even surpasses the state-of-the-art (SOTA) video LLM on long video understanding (\eg, +\textbf{4.6} on the MLVU benchmark). 
We further provide an in-depth ablation study, offering insights into the redundancy of visual tokens and the behavior of individual LLM layers, which can guide future research on multi-modal LLMs.
Altogether, these promising results are attributed to the capability of our adaptive inference that can accommodate diverse computational conditions (\eg, \textbf{40-fold} reduction in FLOPs) with a manageable drop in performance.

\smallskip
Our \textbf{contributions} are summarized as follows:
\begin{itemize}
    \item  We introduce a training-free method that enables adaptive inference of pre-trained multi-modal LLMs. It can reduce visual token redundancy and computational demand while preserving the base model’s performance.
    \item Our approach adopts a minimalist design that merges visual tokens before the LLM and prunes visual tokens within the LLM progressively. This design is applicable to various multi-modal LLM models.
    \item Thanks to adaptive inference, our method achieves a substantial reduction in computational cost while preserving the performance of base image and video LLMs, and even outperforms SOTA video LLMs given similar FLOPs.
\end{itemize}

\section{Related Work}
\label{sec:related_work}

\noindent \textbf{Multi-modal LLMs.}
Large language models \cite{chatgpt, gpt4, llama3_2} exhibit strong performance in text understanding and generation tasks, laying the foundation for developing image LLMs~\cite{dai2023instructblip,zhu2023minigpt4,liu2023improvedllava,li2024llava} and their chain-of-thoughts~\cite{MitraCCoT,zhong-2024-beyond,zhang2023multicot}.
Meanwhile, video LLMs have also made progress via video instruction tuning~\cite{Maaz2023VideoChatGPT,lin2023video,zhu2023languagebind, li2024llava, zhang2024videoinstructiontuningsynthetic, Qwen2VL, ye2024mplugowl3longimagesequenceunderstanding}. 
Recent works~\cite{wang2024videollamblongcontextvideounderstanding, faure2024hermestemporalcoherentlongformunderstanding, nguyen2024encodingcontrollingglobalsemantics, weng2024longvlmefficientlongvideo,  wang2024longllavascalingmultimodalllms, korbar2024textconditionedresamplerlongform, zhang2024long} focus on improving video LLMs for long video understanding, addressing the challenge of a large number of video frames. They either adopt Q-former~\cite{li2023blip} or state-space models~\cite{gu2024mambalineartimesequencemodeling} to aggregate the information before feeding visual tokens into the LLMs. Others extend the context window of LLMs~\cite{wei2024visualcontextwindowextension,zhang2024long,hu2024longreciperecipeefficientlong} or design sequence parallelism from a systems perspective~\cite{xue2024longvila} to process the long video sequence. 
Built on top of the pre-trained image and video LLMs, our adaptive inference method aims to reduce token redundancy during inference and adapt to diverse efficiency requirements while maintaining model performance.

\begin{figure*}[t]
    \centering
    \includegraphics[width=0.90\linewidth]{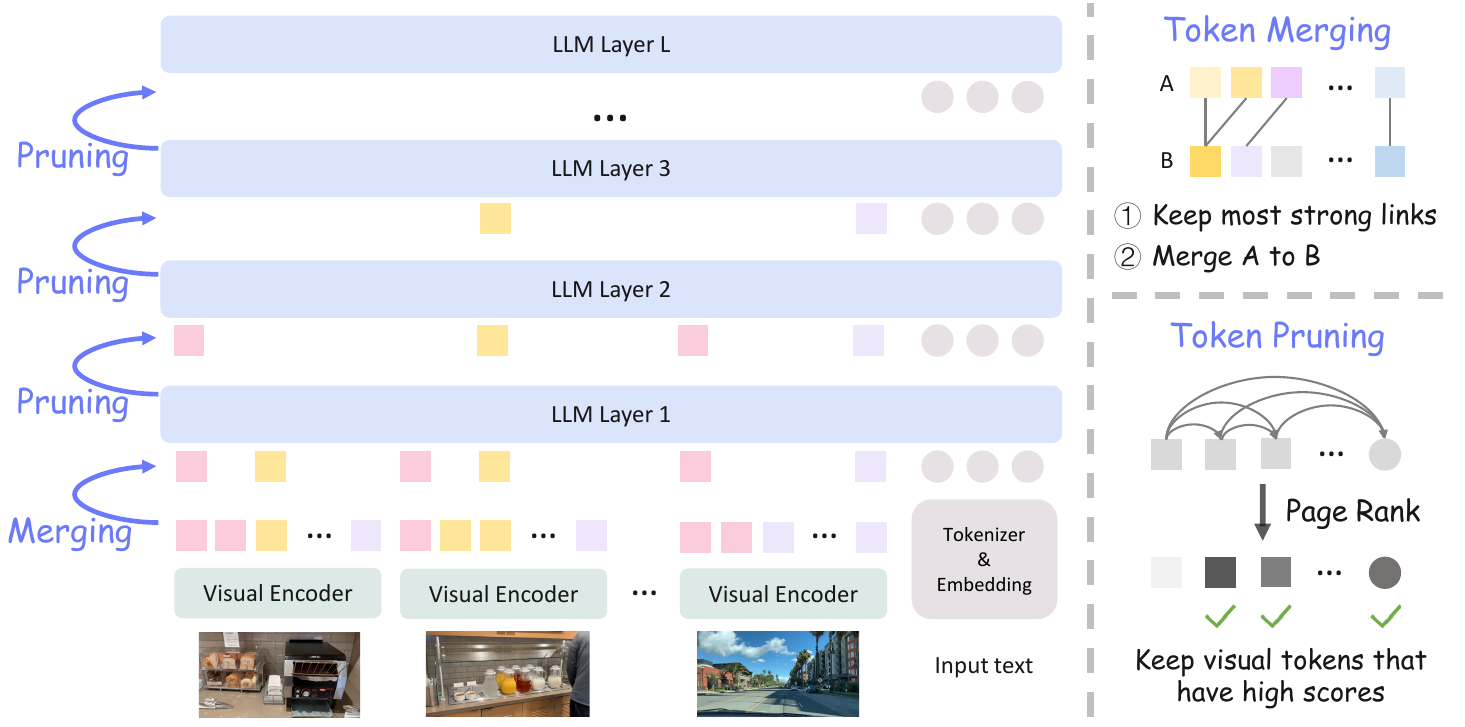}
    \caption{\textbf{Overview} of our training-free method for pre-trained multi-modal LLMs, optimized for accuracy-efficiency trade-off. During inference, we first merge the input visual tokens of LLM based on the cosine similarity between token embeddings, reducing their redundancy. The retained tokens are then fed into the LLM. Token pruning is applied within the LLM layers using the Page Rank algorithm, with a scheduler controlling the retention ratio of each layer. By adjusting the merging ratio and pruning scheduler parameters, our approach enables adaptive inference of multi-modal LLMs, accommodating various computational constraints with minimal performance loss.}
    \label{fig:method}
\end{figure*}

\smallskip
\noindent \textbf{Token Merging and Pruning.} Transformers are widely used in deep learning models, yet they come with high computational demands. Token merging and pruning methods aim to reduce the number of tokens to reduce this load, commonly developed in both NLP models~\cite{powerbert,lat,spatten,trbert,learned} and vision models~\cite{dynamicvit,avit,spvit,tps,savit,dtop,heatvit}. These approaches typically require fine-tuning after token reduction. By contrast, some training-free methods have been developed for efficient vision Transformers~\cite{ats,bolya2022tome,wang2024zero}. Different from these works, we propose a training-free approach tailored for multi-modal LLMs.

For multi-modal LLMs, token pruning has been explored recently~\cite{chen2025image, lin2024boosting, xing2024pyramiddrop, shang2024llava}.
FastV~\cite{chen2025image} and VTW~\cite{lin2024boosting} prune visual tokens at a particular selected LLM layer, while PDrop~\cite{xing2024pyramiddrop} prunes tokens at the end of each stage of LLM layers. LLaVA-Prumerge~\cite{shang2024llava} leverages the key-value pairs from the vision encoder to prune visual tokens before LLM.
Unlike them, our method performs token reduction both before and across LLM layers, allowing for adaptive inference fitting with various computation constraints.

There are concurrent efforts~\cite{zhang2024treat,ye2025fit,zhang2024sparsevlm,tao2025dycoke} that also seek to reduce the computation of multi-modal LLMs. They are only applied to image LLMs, prune tokens exclusively within LLM, or rely on fine-tuning for an accuracy-efficiency balance.
In comparison, our work is training-free, prunes visual tokens before and within LLM, generalizes to both video and image LLMs, and moreover, supports adaptive inference that accommodates a wide range of computation budgets with minimal performance loss.

\smallskip
\noindent \textbf{Adaptive Inference}.
The ability to dynamically adjust the computational complexity of prediction models based on input data, latency constraints, or desired accuracy levels has received considerable interest in the vision and learning community~\cite{han2021dynamic}. Early attempts mainly focus on feature selection within multistage prediction pipelines~\cite{karayev2014anytime,xu2012greedy,grubb2012speedboost}. More recent efforts have extended this concept to the adaptive inference of deep models. Methods have been developed for both convolutional networks and vision Transformers, enabling techniques such as input resampling, operation skipping, layer dropping, and early exiting~\cite{figurnov2017spatially, li20212d, wang2018skipnet, bengio2015conditional, wu2018blockdrop, hu2019learning, jie2019anytime, meng2020ar, wang2021not, dynamicvit, pan2021ia, meng2022adavit,xu2022smartadapt}. Similar ideas have also been explored for LLMs~\cite{du2022glam,rotem2023finding}, and recently extended to MLLMs~\cite{xu2025learning}.
In contrast to these works, we propose an adaptive inference method tailored for multi-modal LLMs via merging similar tokens and pruning unimportant tokens in multi-modal reasoning.


\section{Method}
\label{sec:method}
Multi-modal LLMs for images and videos usually require processing a high volume of visual tokens, resulting in substantial computational costs, especially for video tasks. Such computational load arises from the redundancy within visual tokens.
To tackle this efficiency problem, we propose a training-free adaptive inference method by pruning redundant tokens in two stages, as shown in Figure~\ref{fig:method}. 
First, before visual tokens enter the LLM, we merge highly similar visual tokens, significantly reducing their count while preserving performance. Second, within the LLM, we further rank the visual tokens and remove less important ones at each layer, by applying the Page Rank algorithm to the self-attention weights. 
Together, our approach enables adaptive inference of multi-modal LLMs, reducing computational demands without compromising reasoning performance, and moreover, supports a broad range of computational requirements.

\subsection{Multi-modal LLMs}
Given an image, a typical image LLM~\cite{liu2023improvedllava} first transforms the image into visual tokens via a visual encoder, then projects visual tokens via an adapter (\eg, multilayer perceptron), and finally concatenates the projected visual tokens and text tokens to perform LLM reasoning. 
Video LLMs~\cite{li2024llava} adopt a similar approach to image LLMs, except that the input visual data become uniformly sampled images (video frames), and adaptive pooling is oftentimes applied before LLMs to reduce the number of visual tokens. Despite this pooling, the visual token count remains high (\eg, thousands of tokens per video), resulting in substantial computational demands for the LLM.

Our training-free method is designed to reduce the redundancy of visual tokens, by applying it to the input visual tokens of the LLM and to the visual tokens at each LLM layer. 
We denote the visual tokens right before LLM as $\mathbf {v^0} \in \mathcal{R}^{N^{0} \times D}$, where $N^{0}$ is the initial number of visual tokens.
Within LLM, we denote the input visual tokens and the input text tokens at the $l$-th layer as $\mathbf {v^l} \in \mathcal{R}^{N^{l} \times D}$ and $\mathbf {t^l} \in \mathcal{R}^{M^{l} \times D}$, respectively. $l \in [1, L]$ denotes the index of LLM layer and $N^{l}$ ($M^{l}$) represents the number of visual (text) tokens at layer $l$.


\subsection{Token Merging before LLM}
The input to LLMs often consists of hundreds of visual tokens for an image or thousands for a video, resulting in substantial redundancy. Merging these redundant tokens before they enter the LLM can significantly reduce computational demands. 
Inspired by~\cite{bolya2022tome}, we merge the visual tokens with high similarity, where similarity is quantified by the cosine similarity between token embeddings. 
Unlike ~\cite{bolya2022tome} that merges tokens at each layer of visual encoder, our method perform token merging after visual encoder. This design is agnostic to encoder architectures and easy for plug-and-play.
As illustrated in Figure~\ref{fig:method}, given an initial set of visual tokens $\mathbf {v^0}$ at the first LLM layer, we divide adjacent tokens into sets A and B, calculate pairwise similarity scores between the sets, and identify the closest matching token in set B for each token in set A. We then merge the token pairs with the highest similarity scores by averaging their embeddings. This process reduces the number of tokens by half at most. We repeat this merging process iteratively (\eg, twice) to achieve the desired retention ratio.

For videos, we merge the visual tokens within individual frames. Empirical ablation studies suggest that merging tokens within individual frames has minimal impact on final reasoning performance in video tasks, while merging across frames can be harmful. We hypothesize that merging across frames disrupts temporal order of tokens, leading to a loss of crucial temporal information for video understanding.

\subsection{Token Pruning within LLM}
After merging similar visual tokens, we concatenate the merged visual tokens $\mathbf {v^1}$ with text tokens $\mathbf {t^1}$, forming $\mathbf {x^1} = [\mathbf {v^1}; \mathbf {t^1}]$, which is then passed to the LLM. At each LLM layer, we retain important tokens and prune the less important ones, based on the attention weights computed at each Transformer layer. Specifically, we compute the importance score of each token by applying the Page Rank algorithm~\cite{ilprints422,wang2024zero}, using the attention weights as the adjacency matrix. The importance score $s^l_i$ of token $x^l_i$ at layer $l$ is computed as follows:
\begin{equation}
s^l_i = \frac{1}{N^l + M^l} \sum_{j=1}^{N^l + M^l} \mathbf{A}^l_{i, j} \cdot s^l_j
\end{equation}
where $s^l_j$ is initialized uniformly over all tokens and $\mathbf{A}^l$ represents the softmax-normalized attention weights.
Based on these importance scores, we only prune visual tokens and retain the most important visual tokens, leaving text tokens intact. Our rationale is that pruning text tokens substantially degrades performance, likely because multi-modal LLMs rely on text tokens to perform text-centric reasoning.
As a result, the number of visual tokens input to the next layer is $N^1 \times r^l$, where $r^l$ is the retention ratio at layer $l$.

Further, we design a scheduler to control the retention ratio $r^l$ at the $l$-th layer. It determines the number of visual tokens retained at each LLM layer. Specifically, it is designed as a piece-wise function:

\begin{equation}
r^l = 
\small
\begin{cases} 
1, & \text{if } l < l_1 \\
1 - k (l - l_1), & \text{if } l_1 \leq l \leq l_2 \\
0, & \text{if } l > l_2
\end{cases}
\end{equation}
where $l$ denotes the layer index and $k = \frac{1}{l_2 - l_1}$ represents the slope. In this function, $l_1$ determines from which layer the token pruning starts, $l_2$ defines the layer where the visual tokens are pruned at all, and the difference of $l_1$ and $l_2$ controls the pruning progress. By adjusting the parameters $l_1$ and $l_2$, our approach allows a flexible balance between reasoning accuracy and computational efficiency.

This scheduler design is supported by our empirical findings: pruning visual tokens at the early layers negatively impacts the performance, while in contrast, pruning a large proportion of visual tokens at later layers can still maintain performance. We hypothesize that the LLM prioritizes cross-modal fusion in early layers and shifts focus toward text reasoning in later layers.

\subsection{Adaptive Inference}
By combining token merging and token pruning, our method enables adaptive inference that can meet diverse computational demands. 
Specifically, we can vary the retention ratio of token merging and the scheduler parameters ($l_1$ \& $l_2$) of token pruning, to create a broad range of accuracy-efficiency trade-offs. The specific inference configuration can be determined by the computational requirements in real use cases (\eg, FLOPs and prefill time).
We demonstrate adaptive inference in our experiments and showcase that our method achieves considerable computation reduction while preserving accuracy.


\begin{table*}[t!]
\centering
\setlength{\tabcolsep}{8pt}
\renewcommand{\arraystretch}{1.4}
\resizebox{0.85\textwidth}{!}
{%
\begin{tabular}{@{}l|cc|cccccc@{}}
    \toprule
    \multirow{2}{*}{\textbf{Model}} & 
    \multirow{2}{*}{\textbf{FLOPs}} & 
    \multirow{2}{*}{\textbf{Prefill Time}} &
    {\textbf{\footnotesize{VideoMME}}} &
    {\textbf{\footnotesize{MVBench}}} & 
    {\textbf{\footnotesize{MLVU}}} & 
    {\textbf{\footnotesize{EgoSchema}}} & 
    {\textbf{\footnotesize{NextQA}}} & 
    {\textbf{\footnotesize{PerceptionTest}}}  \\ 
    \cmidrule(l){4-9} 
    & (TB) & (ms) & wo / w-subs & test & m-avg & test & mc & val  \\ \midrule
    \rowcolor{gray!15}\multicolumn{9}{c}{\textbf{Video LLMs}} \\
    LongVA-7B~\cite{zhang2024long} & 381.09 & 2186.04 &  52.6 / 54.3 & -  & 56.3  & -  &  68.3 & - \\
    LLaVA-OV-7B~\cite{li2024llava} & 99.63 & 439.58 & 58.2 / 61.5 & 56.7 & 64.7 & 60.1 & 79.4 & 57.1  \\ \midrule 
    \rowcolor{gray!15}\multicolumn{9}{c}{\textbf{Training-free Method Applied during Inference}} \\
    VTW~\cite{lin2024boosting} & 22.38 & 101.93 & 41.0 / 50.0  &  44.3 &  39.6 &  38.0 & 52.1 & 41.3 \\
    PDrop~\cite{xing2024pyramiddrop} & 24.22 & 104.88 & 51.7 / 56.6 & 52.3 & 55.6 & 51.8 & 74.2 & 52.8 \\
    FastV~\cite{chen2025image}  & 21.24 & 79.56  & 55.9 / 60.0  &  55.9   &  61.1  &  57.5  & 77.5  &  \textbf{56.3}    \\
    LLaVA-Prumerge~\cite{shang2024llava}  & 23.65 &  86.89 & 57.0 / 59.9 & 56.5   &  60.6  &  \textbf{61.0}  &  77.6 &  55.8   \\
    Ours & \textbf{14.76} & \textbf{55.03}  & \textbf{58.2} / \textbf{61.3} & \textbf{57.1}   & \textbf{63.7}  &  59.6  &  \textbf{78.4} &  56.0 \\
    \bottomrule
    \end{tabular}%
}
\vspace{-2mm}
\caption{\textbf{Results on video benchmarks}. Compared to the base model LLaVA-OV-7B, our method significantly reduces FLOPs and prefill time by a factor of \textbf{6.8} and \textbf{8.0}, respectively, with minimum or even without performance loss. Compared to baseline methods, our method outperforms them on most benchmarks with only \textbf{69.5}\% FLOPs and \textbf{69.2}\% prefill time.}
\label{tab:video_bench}
\end{table*}

\begin{table*}[t!]
\centering
\setlength{\tabcolsep}{8pt}
\renewcommand{\arraystretch}{1.4}
\resizebox{0.68\textwidth}{!}
{%
\begin{tabular}{@{}l|ccc|ccc@{}}
    \toprule
    \multirow{2}{*}{\textbf{Model}} & 
    \multirow{2}{*}{\textbf{Number of}} & 
    \multirow{2}{*}{\textbf{FLOPs}} &
    \multirow{2}{*}{\textbf{Prefill Time}} & 
    {\textbf{\footnotesize{VideoMME}}} &
    {\textbf{\footnotesize{MLVU}}} & 
    {\textbf{\footnotesize{EgoSchema}}} \\ 
    \cmidrule(l){5-7} 
    & \textbf{Frames} & (TB) & (ms) & wo / w-subs & m-avg & test \\ \midrule
    \rowcolor{gray!15}\multicolumn{7}{c}{\textbf{Video LLMs}} \\
    LLaVA-OV-7B~\cite{li2024llava} & 32 & 99.63 & 439.58  & 58.2 / 61.5 & 64.7 & 60.1   \\ \midrule  
    \rowcolor{gray!15}\multicolumn{7}{c}{\textbf{Training-free Method Applied during Inference}} \\
    Ours & 32 & \textbf{14.76} & \textbf{55.03} & 58.2 / 61.3 & 63.7  &  59.6  \\
    Ours & 192 & 99.27 & 471.20 & \textbf{59.2} / \textbf{62.3} &  \textbf{69.3}  &  \textbf{60.8}  \\
    \bottomrule
    \end{tabular}%
}
\vspace{-2mm}
\caption{\textbf{Results on long video benchmarks} using more sampled frames as inputs. With comparable FLOPs and prefill time, our method can accommodate more sampled frames and thus improve the base model LLaVA-OV-7B for long video understanding.}
\label{tab:dense_frames}
\end{table*}

\section{Experiments}
\label{sec:experiments}

In this section, we first introduce our implementation details and benchmarks, and then present our results, including the results of video benchmarks, image benchmarks, ablation studies, computation overhead, and adaptive inference.

\smallskip
\noindent\textbf{Implementation Details.}
Our method is applied during the inference of pre-trained multi-modal LLMs.
For video LLMs, we follow the hyperparamers as base model LLaVA-OV-7B~\cite{li2024llava}, such as sampling 32 frames per video unless otherwise noted. It uses Qwen2~\cite{yang2024qwen2technicalreport} as LLM with 28 layers in total. 
For our method, we set the retention ratio of token merging as 25\%, $l_1$ as 14, and $l_2$ as 22.
For image LLMs, we follow the hyperparamers as base model LLaVA-1.5-7B~\cite{liu2023improvedllava} which adopts Vicuna~\cite{vicuna2023} as LLM with 32 layers in total. 
For our method, we set the retention ratio of token merging as 12.5\%, $l_1$ as 13, and $l_2$ as 21. 
We compute FLOPs and prefill time of LLMs using the library from LLM-Viewer~\cite{yuan2024llm}, and assume 100 (40) text tokens for video LLMs (image LLMs).

\smallskip
\noindent\textbf{Benchmarks.}
We evaluate our method on both video and image benchmarks. 
For video LLMs, we consider the following widely adopted benchmarks. VideoMME~\cite{fu2024video} is a comprehensive video benchmark with diverse video durations and video domains. MLVU~\cite{zhou2024mlvucomprehensivebenchmarkmultitask} highlights the reasoning for long videos. Egoschema~\cite{mangalam2023egoschemadiagnosticbenchmarklongform} focuses on ego-centric video understanding. MVBench~\cite{li2024mvbenchcomprehensivemultimodalvideo} and NextQA~\cite{xiao2021nextqanextphasequestionansweringexplaining} focus on temporal action understanding. PercetionTest~\cite{patraucean2023perception} is a comprehensive video benchmark that evaluates perception and reasoning skills.
For image LLMs, we report results on 7 benchmarks. GQA~\cite{hudson2019gqa} and VQA-v2~\cite{goyal2017makingvvqamatter} are classic VQA datasets that assess visual reasoning ability, with GQA placing a particular emphasis on visual attributes. MME~\cite{fu2024mmecomprehensiveevaluationbenchmark} serves as a broad benchmark for evaluating both perception and cognitive abilities. TextVQA~\cite{singh2019vqamodelsread} specifically measures OCR reasoning skills. SQA-IMG~\cite{lu2022learnexplainmultimodalreasoning} covers questions across topics in natural sciences, language sciences, and social sciences. MMB~\cite{liu2024mmbenchmultimodalmodelallaround} tests perception and reasoning capabilities, while POPE~\cite{li2023evaluatingobjecthallucinationlarge} addresses the problem of object hallucination.

\subsection{Video Benchmarks}
Table~\ref{tab:video_bench} and Table~\ref{tab:dense_frames} show the results of our method applied to base video LLM and our method accepting more sampled video frames, respectively.

\begin{table*}[t!]
\centering
\setlength{\tabcolsep}{8pt}
\renewcommand{\arraystretch}{1.4}
\resizebox{0.88\textwidth}{!}
{%
\begin{tabular}{@{}l|cc|ccccccc@{}}
    \toprule
    \multirow{2}{*}{\textbf{Model}} & 
    \multirow{2}{*}{\textbf{FLOPs}} & 
    \multirow{2}{*}{\textbf{Prefill Time}} &
    {\textbf{\footnotesize{VQA-v2}}} & 
    {\textbf{\footnotesize{GQA}}} & 
    {\textbf{\footnotesize{MME}}} & 
    {\textbf{\footnotesize{TextVQA}}} & 
    {\textbf{\footnotesize{SQA-IMG}}} & 
    {\textbf{\footnotesize{MMB}}}  & 
    {\textbf{\footnotesize{POPE}}} \\
    & (TB) & (ms) & (107,394) & (12,578) & (2,374) & (5,000)  &  (2,017) &  (4,377) & (8,910) \\ \midrule
    \rowcolor{gray!15}\multicolumn{10}{c}{\textbf{Image LLMs}} \\
    Qwen-VL-Chat-7B~\cite{bai2023qwen} &  6.44  &  22.51   &  78.2 &  57.5  & 1487.5  & 61.5  &  68.2  & 60.6  &  - \\
    LLaVA-1.5-7B~\cite{liu2023improvedllava}  & 8.18 &  29.30  & 78.5 & 62.0 & 1510.7 & 58.2  &  66.8 &  73.7 & 85.9 \\
    \midrule  
    \rowcolor{gray!15}\multicolumn{10}{c}{\textbf{Training-free Method Applied during Inference}} \\
    VTW~\cite{lin2024boosting} & 2.43 & 13.88 & 49.4  & 42.5  & 916.4 & 45.7 &  66.1 & 63.1 & 17.9 \\
    PDrop~\cite{xing2024pyramiddrop} & 2.36 & 13.31 & 58.1 & 47.3 & 999.0 & 50.4 & \textbf{68.7} & 63.5 & 46.6 \\
    FastV~\cite{chen2025image}  & 2.58 &  10.34  & 74.1   & 56.6  &  1438.5  &  \textbf{57.3}  &  68.0  &  72.1  &  73.6 \\
    LLaVA-Prumerge+~\cite{shang2024llava}  & 2.41 &  9.73  &  74.6  & 57.4  & 1391.9  &  55.2  &  67.9  &  71.6  &  82.2  \\
    Ours & 2.22 & 10.92 &  \textbf{75.4}  & \textbf{58.6}  &  \textbf{1443.5}  &  53.8  &  68.4  &  \textbf{72.5}  & \textbf{85.7}  \\ \midrule
    VTW~\cite{lin2024boosting} & 1.24 & 10.66 & 42.3  & 38.9  & 683.7 &  43.0 & 65.6 & 36.5 & 25.2 \\
    FastV~\cite{chen2025image}  & 1.12 &  9.56  &  55.4   & 45.5  &  960.4 &  51.3  &  66.0  &  61.5 & 33.4  \\
    LLaVA-Prumerge~\cite{shang2024llava}  & 1.04 &  8.99  & 66.7  & 51.3 & 1242.5 & \textbf{53.8} &  \textbf{68.0} & 67.1 & 76.2 \\
    Ours & 1.00 & 8.98 & \textbf{69.0} & \textbf{54.6} &  \textbf{1277.7} &  48.4  &  67.1 & \textbf{69.4} & \textbf{79.5} \\ 
    \bottomrule
    \end{tabular}%
}
\vspace{-2mm}
\caption{\textbf{Results on image benchmarks}. VQA-v2 and GQA are the most stable benchmarks that have the most evaluation samples. 
With less computation cost, our method outperforms baselines on most benchmarks.
Note that PDrop keeps all tokens at the first 25\% LLM layers and thus does not support low FLOPs, \ie, below 25\% FLOPs of base model.
Compared to the base model LLaVA-1.5-7B, our model significantly reduces FLOPs and prefill time (\eg, by a factor of \textbf{3.7} and \textbf{2.7}, respectively), with a manageable performance loss. 
}
\label{tab:image_bench}
\end{table*}

\smallskip
\noindent\textbf{Setup.} We choose LLaVA-OV-7B~\cite{li2024llava} as our base model and follow its evaluation protocol. Specifically, during inference of the pre-trained base model, we apply our token merging to the input visual tokens of Qwen2 LLM and perform token pruning at layers of Qwen2. We measure our model performance on diverse video benchmarks and report efficiency metrics (FLOPs and prefill time) and accuracy metrics of each video benchmark.

\smallskip
\noindent \textbf{Baselines.} 
Our baselines include (1) FastV~\cite{chen2025image} that prunes visual tokens at a particular LLM layer, (2) VTW~\cite{lin2024boosting} which abandons all visual tokens at a particular selected LLM layer, (3) PDrop~\cite{xing2024pyramiddrop} that divides LLM layers into four stages and prunes tokens at the end of each stage, based on the attention scores between visual tokens and the last text token, and (4) LLaVA-1.5-Prumerge~\cite{shang2024llava} that leverages the key-query pairs in the visual encoder to prune and merge visual tokens. 
We conduct the experiments with the pre-trained LLaVA-OV-7B model in the training-free setting.

\smallskip
\noindent\textbf{Accuracy-efficiency balance.}
As Table~\ref{tab:video_bench} shows, our model achieves substantial reductions in computational demands compared to the base model LLaVA-OV-7b (\eg, 14.76 vs. 99.63 FLOPs), decreasing FLOPs by a factor of \textbf{6.8} and prefill time by \textbf{8.0}, with little or no performance drop across diverse benchmarks. 
Additionally, compared to the baseline methods, our model consistently achieves higher performance on most benchmarks, while only requiring \textbf{69.5}\% FLOPs and \textbf{69.2}\% prefill time at most. These results indicate that our method effectively reduces redundancy in visual tokens, providing an optimal balance between accuracy and efficiency.

\smallskip
\noindent\textbf{More sampled frames improve long video understating with close efficiency.}
In Table~\ref{tab:dense_frames}, our model significantly reduces the FLOPs and prefill time of the base pre-trained model, enabling the sampling of more frames in videos (\eg, from 32 to 192 frames). With comparable total computation (\ie, 99.27 vs. 99.63 FLOPs), our model improves performance on long video understanding benchmarks, especially on long video benchmark MLVU, with a notable gain of +\textbf{4.6}. This improvement is attributed to our method’s ability to retain essential visual tokens with much less redundancy, while densely sampled frames capture additional information crucial for long video comprehension.

\subsection{Image Benchmarks}
The results are reported in Table~\ref{tab:image_bench}, comparing our model with the base model and baseline methods.

\smallskip
\noindent\textbf{Setup.} We choose LLaVA-1.5-7B~\cite{liu2023improvedllava} as our base model following its evaluation protocol. Again, during inference of the pre-trained base model, we merge the input visual tokens of Vicuna LLM and prune visual tokens at Vicuna's layers.  We test our model on several image benchmarks and report efficiency metrics (FLOPs and prefill time) and accuracy metrics of each image benchmark.  

\smallskip
\noindent \textbf{Baselines.} 
We again choose the baselines as FastV~\cite{chen2025image}, VTW~\cite{lin2024boosting}, PDrop~\cite{xing2024pyramiddrop}, and LLaVA-1.5-Prumerge~\cite{shang2024llava} which provide training-free results on image benchmarks. We also compare with a variant LLaVA-1.5-Prumerge+ which trades computation for reasoning performance. Both our models and baselines are evaluated in the training-free setting with LLaVA-1.5-7B as the base model.

\smallskip
\noindent\textbf{Adaptive inference compared with the base model.} As shown in Table~\ref{tab:image_bench}, when compared to the base model LLaVA-1.5-7B, our model (with 1.00 FLOPs) significantly reduces the FLOPs and prefill time (\ie, using only \textbf{12.5}\% FLOPS and \textbf{30.6}\% prefill time), with a noticeable performance drop. 
However, by adjusting the merging ratio and pruning scheduler, our model (with 2.22 FLOPs) largely mitigates this performance gap while still considerably reducing FLOPs and prefill time (\ie, using only \textbf{27.1}\% FLOPS and \textbf{37.3}\% prefill time). This capability is supported by our design of adaptive inference that can accommodate various efficiency demands.

\smallskip
\noindent\textbf{Comparison with baseline methods.}
In Table~\ref{tab:image_bench}, with the same level of FLOPs and prefill time, our model consistently outperforms the baseline methods on most benchmarks by a clear margin (\eg, +2.3 on VQA-v2, +3.3 on GQA, +33.2 on MME, +2.3 on MMB, +3.3 on POPE over LLaVA-Prumerge). These results suggest that our method can effectively retain visual tokens critical for image reasoning tasks, while reducing computation costs.

We also notice that our model performs less satisfactorily on TextVQA which includes text-intensive images and requires a model to heavily preserve the textual information. We conjecture that LLaVA-Prumerge leverages the self-attention weights from the visual encoder, while FastV retains tokens that are important to text tokens, thereby being more friendly to text-rich tasks.

\subsection{Ablation Study}
In this section, we conduct ablation studies for our method, by adjusting the retention ratio of token merging, the parameters of the token pruning scheduler, and the pruning strategy. Their results are presented in Table~\ref{tab:ablation_merging}, Table~\ref{tab:ablation_pruning}, and Table~\ref{tab:ablation_text}, respectively. More ablations are in the Appendices.

\begin{table}[t!]
\centering
\setlength{\tabcolsep}{8pt}
\renewcommand{\arraystretch}{1.2}
\resizebox{0.35\textwidth}{!}
{%
\begin{tabular}{@{}l|cc|c@{}}
    \toprule
    \multirow{2}{*}{\textbf{Retention}} & 
    \multirow{2}{*}{\textbf{FLOPs}} &
    \multirow{2}{*}{\textbf{Prefill Time}} & 
    {\textbf{\footnotesize{VideoMME}}}  \\ 
    \cmidrule(l){4-4} 
    \textbf{Ratio} & (TB) & (ms) & wo-subs \\ \midrule
    100.0\% & 99.63 & 439.58  & 58.2   \\ \midrule  
    50.0\%  & 46.48 & 182.65  & 58.5  \\
    \textbf{25.0\%}  & \textbf{22.90} & \textbf{83.94}   & \textbf{58.0}  \\
    12.5\%  & 11.64 & 41.22   & 56.6  \\
    6.3\%   & 6.41  & 22.54   & 53.6  \\
    3.1\%   & 3.85  & 13.68   & 52.3  \\
    1.6\%   & 2.57  & 10.15   & 50.9  \\
    \bottomrule
    \end{tabular}%
}
\vspace{-2mm}
\caption{\textbf{Ablation study on token merging}. We show the performance on the VideoMME benchmark by varying various retention ratios of token merging, with token pruning in LLMs disabled. }
\label{tab:ablation_merging}
\end{table}

\begin{table}[t!]
\centering
\setlength{\tabcolsep}{8pt}
\renewcommand{\arraystretch}{1.2}
\resizebox{0.40\textwidth}{!}
{%
\begin{tabular}{@{}l|cc|cc|c@{}}
    \toprule
    \multirow{2}{*}{\textbf{Exp.}} & 
    \multirow{2}{*}{\textbf{$l_1$}} &
    \multirow{2}{*}{\textbf{$l_2$}} & 
    \multirow{2}{*}{\textbf{FLOPs}} &
    \multirow{2}{*}{\textbf{Prefill Time}} & 
    {\textbf{\footnotesize{VideoMME}}}  \\ 
    \cmidrule(l){6-6} 
     & & & (TB) & (ms) & wo-subs \\ \midrule
    1 & 28 & 29 & 22.90 & 83.94  & 58.0   \\   
    2 & 21 & 29 & 20.15 & 73.61  & 58.0  \\
    3 & 14 & 29 & 17.41 & 63.34  & 57.7 \\
    4 & 7  & 29 & 14.66 & 53.08  & 57.4  \\ \midrule
    5 & 21 & 22 & 17.50 & 65.35  & 58.1  \\
    6 & \textbf{14} & \textbf{22} & \textbf{14.76} & \textbf{55.03}  & \textbf{58.2}  \\
    7 & 7  & 22 & 12.01 & 44.75  & 56.8  \\ \midrule
    8 & 14 & 15 & 12.10 & 46.77  & 54.3  \\
    9 & 7  & 15 & 9.36  & 36.44  & 52.9  \\ \midrule
    10& 7  & 8  & 6.71  & 28.18  & 41.9  \\
    \bottomrule
    \end{tabular}%
}
\vspace{-2mm}
\caption{\textbf{Ablation study on token pruning}. We show the performance on VideoMME by varying parameters of our pruning scheduler ($l_1$ \& $l_2$), with 25\% retention ratio for token merging. }
\label{tab:ablation_pruning}
\end{table}

\smallskip
\noindent\textbf{Setup.} We perform experiments on the VideoMME benchmark as it evaluates models on the videos with various durations and diverse domains.
Specifically, we vary the retention ratio of token merging (Table~\ref{tab:ablation_merging}), scheduler parameters in token pruning (Table~\ref{tab:ablation_pruning}), and the strategy whether prune text tokens or not (Table~\ref{tab:ablation_text}), respectively. 
For token merging, we disable token pruning and investigate the effects of altering the retention ratios.
For token pruning, we set the retention ratio of token merging as 25\% and study the effects of various pruning schedulers or strategies.
Following our main experiments, the metrics include the efficiency aspect (FLOPs and prefill time) and the accuracy aspect.

\smallskip
\noindent\textbf{Token merging substantially reduces redundancy.}
As shown in Table~\ref{tab:ablation_merging}, video reasoning performance remains stable when the retention ratio is set at 25\% or higher, while FLOPs and prefill time are significantly reduced (\eg, to only \textbf{23}\% of FLOPs and \textbf{19}\% of prefill time relative to the base model). 
This suggests that most visual tokens are redundant, with only around one-quarter providing essential information for video understanding tasks.
Moving forward, as the retention ratio is further reduced, there is a gradual decline in reasoning performance, accompanied by substantial savings in FLOPs and prefill time (\eg, down to \textbf{3}\% FLOPs and \textbf{2}\% prefill time of the base model). 
These findings indicate that with retention ratios below 25\%, the model will trade accuracy for efficiency, making it suitable for scenarios requiring high efficiency with less accuracy demands, such as mobile devices.

\smallskip
\noindent\textbf{Controlling accuracy-efficiency trade-offs with token pruning scheduler.} 
As shown in Table~\ref{tab:ablation_pruning}, by adjusting the pruning scheduler parameters $l_1$ and $l_2$, our method enables a broad range of accuracy-efficiency trade-offs. FLOPs decrease from 22.90 to 6.71, and prefill time drops from 83.94 to 28.18, with performance unaffected until FLOPs and prefill time are reduced to approximately half.
Furthermore, in each block, when comparing the first row with subsequent rows, we observe a reduction in FLOPs and prefill time with little to no impact on performance. For example, Exp. 6 achieves a 16\% reduction in both FLOPs and prefill time compared to Exp. 5, while maintaining performance. These findings validate our token pruning design, demonstrating its flexibility in controlling LLM computations and achieving optimal accuracy-efficiency trade-offs.

\smallskip
\noindent\textbf{Early LLM layers vs.\ later LLM layers.} We find that LLM emphasizes multi-modal fusion at early layers and shifts the focus to text-centric reasoning at later layers. 
For example, experiments 5, 8, and 10 in Table~\ref{tab:ablation_pruning} remove all visual tokens after layer 22, layer 15, and layer 8, respectively. Performance is maintained when visual tokens are removed starting from layer 22 (Exp. 5), but it declines when tokens are removed from layer 15 (Exp. 8) and more significantly from layer 8 (Exp. 10). 
Notably, removing tokens as early as layer 8 causes a substantial performance drop (\eg, from 58.0 to 41.9). 
Based on these findings, we adopt a strategy (Exp. 6) that retains visual tokens in early layers, gradually prunes them in middle layers, and completely removes them in later layers.

\smallskip
\noindent\textbf{Text tokens matter in LLM layers.} In Table~\ref{tab:ablation_text}, involving text tokens in the pruning process results in a substantial performance drop (\eg, from 58.2 to 45.7). This finding aligns with the understanding that LLMs primarily perform text-based reasoning, making it essential to retain text tokens throughout inference.

\begin{table}[t!]
\centering
\setlength{\tabcolsep}{8pt}
\renewcommand{\arraystretch}{1.2}
\resizebox{0.42\textwidth}{!}
{%
\begin{tabular}{@{}c|cc|c@{}}
    \toprule
    {\textbf{Retention Ratio}} & 
    {\textbf{($l_1$,$l_2$)}} &
    {\textbf{Prune Text Tokens}} & 
    {\textbf{\footnotesize{VideoMME}}}  \\ \midrule
    \textbf{25\%}  & \textbf{(14,22)} & \textbf{$\times$}  & \textbf{58.2}   \\ 
    25\%  & (14,22) & $\checkmark$ & 45.7  \\
    \bottomrule
    \end{tabular}%
}
\vspace{-2mm}
\caption{\textbf{Ablation study on pruning text tokens within LLM}. We report the performance on VideoMME by comparing our method with and without text token pruning.}
\label{tab:ablation_text}
\end{table}

\smallskip
\noindent \textbf{Overhead of our method.} 
As in Table~\ref{tab:flops_ours}, compared to the FLOPs of LLM inference, the additional FLOPs introduced by our method are minimal---only 0.6\% FLOPs of Qwen2 and 0.03\% FLOPs of Vicuna-v1.5. 
These results show that our overhead (cost) is negligible in comparison to the multi-fold FLOPs reduction (benefit).

\begin{table}[t!]
\centering
\setlength{\tabcolsep}{8pt}
\renewcommand{\arraystretch}{1.4}
\resizebox{0.35\textwidth}{!}
{%
\begin{tabular}{@{}l|cc@{}}
\toprule
\textbf{FLOPs}         & \textbf{Video LLM}  & \textbf{Image LLM}        \\
(GB)          & (Qwen2-7B) & (Vicuna-v1.5-7B) \\ \midrule
Token Merging & 88.25      &   0.23           \\
Token Pruning & 4.18       &   0.03           \\ 
Total       & 92.43      &   0.26               \\ \midrule
LLM Inference & 14757      &   1003         \\ \bottomrule
\end{tabular}
}
\vspace{-2mm}
\caption{\textbf{Computational cost in GFLOPs} introduced by our method. The cost is much less than the FLOPs of LLM inference.}
\label{tab:flops_ours}
\end{table}

\subsection{Adaptive Inference}
As shown in Table~\ref{tab:adaptive}, by combining token merging and token pruning, our method supports a broad range of accuracy-efficiency trade-offs. 
For instance, compared to the base model (99.63 FLOPs and 58.2 accuracy), our default configuration achieves the same accuracy with significantly reduced FLOPs (14.76), while an efficient configuration reduces FLOPs even further (2.51) with an acceptable performance decrease (50.9 accuracy). Overall, our approach spans a \textbf{40-fold} reduction in FLOPs with less than a 13\% drop in accuracy.
This flexibility is enabled by adjusting the key parameters, including the retention ratio in token merging and $l_1$ \& $l_2$ in token pruning. These configurations make our adaptive inference method suitable for various devices and efficiency requirements, such as AR glasses, mobile phones, personal computers, and robots. More results can be found in the Appendices.

\begin{table}[t!]
\centering
\setlength{\tabcolsep}{8pt}
\renewcommand{\arraystretch}{1.2}
\resizebox{0.40\textwidth}{!}
{%
\begin{tabular}{@{}l|cc|cc|c@{}}
    \toprule
    \multirow{2}{*}{\textbf{Retention}} & 
    \multirow{2}{*}{\textbf{$l_1$}} &
    \multirow{2}{*}{\textbf{$l_2$}} & 
    \multirow{2}{*}{\textbf{FLOPs}} &
    \multirow{2}{*}{\textbf{Prefill Time}} & 
    {\textbf{\footnotesize{VideoMME}}}  \\ 
    \cmidrule(l){6-6} 
    Ratio & & & (TB) & (ms) & wo-subs \\ \midrule
    100.0\% & - & - & 99.63 & 439.58  & 58.2   \\   \midrule
    50.0\% & - & - & 46.48 & 182.65  & 58.5  \\
    \textbf{25.0\%} & \textbf{14} & \textbf{22} & \textbf{14.76} & \textbf{55.03}  & \textbf{58.2} \\
    12.5\% &  14 & 22 & 11.14 & 39.41  & 56.4  \\ 
    6.3\% & 14 & 22 & 6.17 & 21.69  & 53.6  \\
    3.1\% & 14 & 22 & 3.72 & 13.26 & 52.3  \\
    1.6\% & 14 & 22 & 2.51 & 10.12 & 50.9  \\ 
    \bottomrule
    \end{tabular}%
}
\vspace{-2mm}
\caption{\textbf{Adaptive inference}. By varying the retention ratio of token merging and the parameters of the pruning scheduler ($l_1$ \& $l_2$), our method supports a broad range of computational conditions without or with a manageable performance drop.}
\label{tab:adaptive}
\end{table}

\section{Conclusion}
In this work, we present a training-free approach of adaptive inference for multi-modal LLMs, through token merging based on visual similarity and token pruning based on multi-modal importance. 
Extensive experiments demonstrate that our method significantly reduces computational demands while maintaining reasoning performance, such as a 6.8/3.7-fold reduction in FLOPs across video/image benchmarks and improved SOTA performance on long video benchmarks. 
Additionally, our key findings reveal that only a small fraction of visual tokens are necessary for multi-modal understanding, and progressive token pruning across LLM layers further optimizes computational efficiency.
We hope that our work will provide a foundation for future advancements in adaptive multi-modal LLMs, in which the accuracy-efficiency tradeoffs can be dynamically adjusted in response to varying computing environments.

\smallskip 
\noindent\textbf{Acknowledgements.} 
This work was supported by National Key R\&D Program of China (Project No.~2022ZD0161200, 2022ZD0161201), by Hong Kong Research Grant Council - Early Career Scheme (Grant No.~24200223), and by Hong Kong Innovation and Technology Commission Project No.~ITS/228/22FP.~ Z.\ L.\ and Y.\ L.\ were not supported by the aforementioned grants.

{
    \small
    \bibliographystyle{ieeenat_fullname}
    \bibliography{main}
}

\clearpage

\setcounter{figure}{0}
\setcounter{table}{0}
\setcounter{section}{0}
\renewcommand{\theequation}{\Alph{equation}}
\renewcommand{\thefigure}{\Alph{figure}}
\renewcommand{\thesection}{\Alph{section}}
\renewcommand{\thetable}{\Alph{table}}

\appendix
\addappheadtotoc
\begin{appendices}

In the appendices, we provide more detailed results in addition to our main paper, including additional ablation study, additional discussion on FlashAttention, and additional results on video benchmarks.

\section{Addtional Ablation Study}

\smallskip
\noindent \textbf{Ablation study for additional frames.} As shown in Table~\ref{tab:ablation_moreframes}, dense frames can improve the performance of VideoMME and MLVU. In comparison, our method not only reduces computation cost significantly, but also creates a token sequence with less redundancy, which in return further improves performance across all three benchmarks.

\smallskip
\noindent \textbf{Ablation study for token merging across frames.} We conducted additional experiments with temporal merging, where tokens from adjacent frames are merged iteratively.
As shown in Table~\ref{tab:ablation_temporal}, merging across frames leads to degraded performance, especially at lower retention ratios, validating our hypothesis in main paper that temporal merging disrupts the temporal order of tokens, resulting in a negative impact on performance.

\smallskip
\noindent \textbf{Ablation study for base models.} 
We conducted additional experiments with Qwen2-VL-7B-Instruct~\cite{Qwen2VL} and adjusted MAX\_PIXELS to a smaller value due to GPU memory limitations.
While LLaVA-Prumerge~\cite{shang2024llava} is one of our baselines in the paper, it is not compatible with Qwen2-VL. This is because LLaVA-Prumerge assumes the use of an image-level CLS token, whereas Qwen2-VL encodes sampled video frames together and does not have an image-level CLS token. 
Results of FastV and our method applied to LLaVA-OV and Qwen2-VL are reported in Table\ \ref{tab:ablation_qwen2vl}. 
Our method consistently outperforms the baseline while requiring fewer FLOPs and prefill time (same conclusion as Table 1 in paper). 
We also notice that the performance drop with Qwen2-VL is larger than with LLaVA-OV. It is likely due to the video encoder in Qwen2-VL, which mixes features of all frames. As shown in above paragraph, cross-frame token merging may disrupt temporal order and is less effective. 

\section{Additional Discussion on FlashAttention}
Our method explores token merging and pruning for adaptive inference in multi-modal LLMs, a direction that is orthogonal to works on improving LLM efficiency, such as quantization~\cite{dettmers2022gpt3}, sparse attention~\cite{child2019generating}, and efficient attention (\eg FlashAttention~\cite{dao2023flashattention}). 
Notably, our method is compatible with quantization and sparse attention, yet not with optimizations like FlashAttention (FA), where attention values are not explicitly computed. 
This is because, similar to prior work on token pruning~\cite{chen2025image}, our method relies on attention values for selecting tokens. 
In Table~\ref{tab:ablation_flashattn}, we conduct a cost-benefit analysis to compare our token pruning with FA. Our method reduces much more prefill time than FA (e.g., -28.90 ms vs.\ -4.83 ms).
In fact, even though FA improves the efficiency of attention mechanisms, with large token numbers, the computation cost remains high.
Integrating the idea of FA and token pruning might be possible (\eg sequence parallelism, matrix approximation), which we leave as future work.

Further, FA was introduced to reduce memory I/O access and accelerate computation, making it particularly beneficial for model training, where backward propagation demands substantial memory and compute resources. However, during inference, its advantages are less pronounced, and its use becomes optional. Instead, the number of tokens processed plays a more significant role in inference efficiency, as shown in Table~\ref{tab:ablation_flashattn}.

\begin{table}[]  
\centering  
\scalebox{0.75}{
\begin{tabular}{lccc}  
\toprule
       Model  & VideoMME & MLVU &  Egoschema  \\
     \midrule
     \rowcolor{gray!15} LLaVA-OV   &  58.2 & 64.7  & 60.1 \\ 
     LLaVA-OV + 128 frames & 58.4 & 67.7 & 59.8 \\ 
     LLaVA-OV + 128 frames + Ours & \textbf{58.9} & \textbf{69.0} & \textbf{60.5} \\ \bottomrule
\end{tabular}
}
\caption{Ablation study for dense frames on long video understanding.
\label{tab:ablation_moreframes}  
}  
\end{table}

\begin{table}[]  
\centering  
\scalebox{0.78}{
\begin{tabular}{lccccc}  
\toprule
       Retention Ratio  & 50\% & 25\% &  12.5\% & 6.3\% & 3.1\% \\
     \midrule
     Temporal Merging    &  57.9 & 55.8  & 54.5 & 50.4 & 47.4   \\
     Spatial Merging (our default)   & \textbf{58.5}  & \textbf{58.0}  & \textbf{56.6} & \textbf{53.6} & \textbf{52.3} \\  \bottomrule
\end{tabular}
}
\caption{Ablation study for temporal or spatial merging on VideoMME.
\label{tab:ablation_temporal}  
}  
\end{table}

\begin{table}[]  
\centering  
\scalebox{0.63}{
\begin{tabular}{lcccccc}  
\toprule
       Model  & FLOPs & Prefill Time &  VideoMME & MVBench & MLVU & Egoschema \\
     \midrule
     \rowcolor{gray!15} LLaVA-OV    &  99.63 & 439.58  & 58.2 & 56.7 & 64.7 & 60.1   \\
     FastV~\cite{chen2025image}   & 21.24  & 79.56  & 55.9 & 55.9 & 61.1 & 57.5 \\
     Ours & \textbf{14.76} & \textbf{55.03} & \textbf{58.2} & \textbf{57.1} & \textbf{63.7} & \textbf{59.6} \\ \midrule
     \rowcolor{gray!15} Qwen2-VL & 61.90 & 252.88 & 55.2 & 62.6 & 58.2 & 61.5 \\
     FastV~\cite{chen2025image} & 14.07 & 51.11 & 51.2 & 57.7 & 54.2 & 57.7 \\
     Ours & \textbf{9.96} & \textbf{36.76} & \textbf{52.8} & \textbf{62.6} & \textbf{57.2} & \textbf{58.1} \\ \bottomrule
\end{tabular}
}
\caption{Ablation study for different base models.
\label{tab:ablation_qwen2vl}  
}  
\end{table}

\begin{table}[t!]
\centering
\setlength{\tabcolsep}{8pt}
\renewcommand{\arraystretch}{1.4}
\resizebox{0.46\textwidth}{!}
{%
\begin{tabular}{@{}l|cc@{}}
\toprule
 Inference & \textbf{FLOPs} (TB)  & \textbf{Prefill Time} (ms)    \\ \midrule
Token Merging & 22.90      &   83.93          \\ 
Token Merging \& FlashAttention & 22.90       &   79.10        \\ 
Token Merging \& Token Pruning        & 14.76      &   55.03   \\ \bottomrule
\end{tabular}
}
\caption{Ablation of FlashAttention vs. pruning on VideoMME.
\label{tab:ablation_flashattn}  
}
\end{table}

\section{Additional Results on Video Benchmarks}

\begin{table*}[t]
\centering
\setlength{\tabcolsep}{8pt}
\renewcommand{\arraystretch}{1.4}
\resizebox{0.88\textwidth}{!}
{%
\begin{tabular}{@{}l|cc|cccccc@{}}
    \toprule
    \multirow{2}{*}{\textbf{Model}} & 
    \multirow{2}{*}{\textbf{FLOPs}} & 
    \multirow{2}{*}{\textbf{Prefill Time}} &
    {\textbf{\footnotesize{VideoMME}}} &
    {\textbf{\footnotesize{MVBench}}} & 
    {\textbf{\footnotesize{MLVU}}} & 
    {\textbf{\footnotesize{EgoSchema}}} & 
    {\textbf{\footnotesize{NextQA}}} & 
    {\textbf{\footnotesize{PerceptionTest}}}  \\ 
    \cmidrule(l){4-9} 
    & (TB) & (ms) & wo / w-subs & test & m-avg & test & mc & val  \\ \midrule
    \rowcolor{gray!15}\multicolumn{9}{c}{\textbf{Video LLMs}} \\
    LongVA-7B~\cite{zhang2024long} & 381.09 & 2186.04 &  52.6 / 54.3 & -  & 56.3  & -  &  68.3 & - \\
    LLaVA-OV-7B~\cite{li2024llava} & 99.63 & 439.58 & 58.2 / 61.5 & 56.7 & 64.7 & 60.1 & 79.4 & 57.1  \\ \midrule 
    \rowcolor{gray!15}\multicolumn{9}{c}{\textbf{Training-free Method Applied during Inference}} \\
    LLaVA-Prumerge~\cite{shang2024llava}  & 23.65 &  86.89 & 57.0 / 59.9 & 56.5   &  60.6  &  \textbf{61.0}  &  77.6 &  55.8   \\
    Ours & 22.06 & 84.36  & 58.0 / \textbf{61.3} & \textbf{57.3}   & \textbf{64.4}  &  59.8  &  78.3 &  \textbf{56.7} \\
    Ours & \textbf{14.76} & \textbf{55.03}  & \textbf{58.2} / \textbf{61.3} & 57.1   & 63.7  &  59.6  &  \textbf{78.4} &  56.0 \\
    \bottomrule
    \end{tabular}%
}
\caption{Additional results on video benchmarks. Supported by our adaptive inference method, we can adjust the parameters in our method to achieve different accuracy-efficiency balance. In this table, we add one of our model variants that consumes more computation resources while achieving slightly better accuracy than our default model.}
\label{tab:video_bench_additional}
\end{table*}

Our method is characterized by the adaptive inference that can adjust accuracy-efficiency trade-offs based on contextual factors, such as the FLOP budget. Below, we present more results of adaptive inference on video benchmarks by assuming a target FLOP budget.

In Table~\ref{tab:video_bench_additional}, to match the computation cost of baseline method LLaVA-Prumerge (\ie, 23.65 FLOPs), we adjust the parameters of our method and create a model variant with comparable computation demand (\ie, 22.06 FLOPs). 
Despite fewer FLOPs, this model variant again largely outperforms LLaVA-Prumerge across most benchmarks (\eg, +1.0 on VideoMME, +3.8 on MLVU, +0.9 on PerceptionTest). 
Further, compared to our default model, this model variant achieves comparable performance on most benchmarks and slightly better results on others (\eg, +0.7 on MLVU, +0.7 on PerceptionTest). 
These results showcase the flexibility of our adaptive inference method, which can optimize the accuracy-efficiency trade-off to fit with specific contextual requirements.

\cleardoublepage

\end{appendices}

\end{document}